\documentclass{article}

\usepackage[english]{babel}

\usepackage[letterpaper,top=2cm,bottom=2cm,left=3cm,right=3cm,marginparwidth=1.75cm]{geometry}

\usepackage{amsmath}
\usepackage{graphicx}
\usepackage[colorlinks=true, allcolors=blue]{hyperref}
\usepackage{booktabs}
\usepackage{bm}
\usepackage{amsfonts}
\usepackage{tabularx}
\usepackage{multirow}
\usepackage{makecell}
\usepackage{mathtools}
\usepackage{pdfpages}
\usepackage{subcaption}
\usepackage{algorithm}
\usepackage{algorithmic}
\usepackage{amsthm}

\title{Asymmetric Learning for Graph~Neural~Network~based~Link~Prediction}
\author{
	Kai-Lang Yao and Wu-Jun Li\footnote{Corresponding Author}\\
	National Key Laboratory for Novel Software Technology,\\
	Nanjing University, 210023, China\\
	\texttt{yaokl@smail.nju.edu.cn and liwujun@nju.edu.cn} \\
}

\begin{document}
\maketitle

\def\a{{\bf a}}
\def\b{{\bf b}}
\def\bbb{{\bf b}}
\def\c{{\bf c}}
\def\d{{\bf d}}
\def\e{{\bf e}}
\def\f{{\bf f}}
\def\g{{\bf g}}
\def\h{{\bf h}}
\def\i{{\bf i}}
\def\j{{\bf j}}
\def\k{{\bf k}}
\def\l{{\bf l}}
\def\m{{\bf m}}
\def\n{{\bf n}}
\def\o{{\bf o}}
\def\p{{\bf p}}
\def\q{{\bf q}}
\def\r{{\bf r}}
\def\s{{\bf s}}
\def\t{{\bf t}}
\def\u{{\bf u}}
\def\v{{\bf v}}
\def\w{{\bf w}}
\def\x{{\bf x}}
\def\y{{\bf y}}
\def\z{{\bf z}}

\def\A{{\bf A}}
\def\B{{\bf B}}
\def\C{{\bf C}}
\def\D{{\bf D}}
\def\E{{\bf E}}
\def\F{{\bf F}}
\def\G{{\bf G}}
\def\H{{\bf H}}
\def\I{{\bf I}}
\def\J{{\bf J}}
\def\K{{\bf K}}
\def\L{{\bf L}}
\def\M{{\bf M}}
\def\N{{\bf N}}
\def\O{{\bf O}}
\def\P{{\bf P}}
\def\Q{{\bf Q}}
\def\R{{\bf R}}
\def\S{{\bf S}}
\def\T{{\bf T}}
\def\U{{\bf U}}
\def\V{{\bf V}}
\def\W{{\bf W}}
\def\X{{\bf X}}
\def\Y{{\bf Y}}
\def\Z{{\bf Z}}

\def\0{{\bf 0}}
\def\1{{\bf 1}}
\def\2{{\bf 2}}
\def\3{{\bf 3}}
\def\4{{\bf 4}}
\def\5{{\bf 5}}
\def\6{{\bf 6}}
\def\7{{\bf 7}}
\def\8{{\bf 8}}
\def\9{{\bf 9}}

\def\AM{{\mathcal A}}
\def\BM{{\mathcal B}}
\def\CM{{\mathcal C}}
\def\DM{{\mathcal D}}
\def\EM{{\mathcal E}}
\def\FM{{\mathcal F}}
\def\GM{{\mathcal G}}
\def\HM{{\mathcal H}}
\def\IM{{\mathcal I}}
\def\JM{{\mathcal J}}
\def\KM{{\mathcal K}}
\def\LM{{\mathcal L}}
\def\MM{{\mathcal M}}
\def\NM{{\mathcal N}}
\def\OM{{\mathcal O}}
\def\PM{{\mathcal P}}
\def\QM{{\mathcal Q}}
\def\RM{{\mathcal R}}
\def\SM{{\mathcal S}}
\def\TM{{\mathcal T}}
\def\UM{{\mathcal U}}
\def\VM{{\mathcal V}}
\def\WM{{\mathcal W}}
\def\XM{{\mathcal X}}
\def\YM{{\mathcal Y}}
\def\ZM{{\mathcal Z}}

\def\AB{{\mathbb A}}
\def\BB{{\mathbb B}}
\def\CB{{\mathbb C}}
\def\DB{{\mathbb D}}
\def\EB{{\mathbb E}}
\def\FB{{\mathbb F}}
\def\GB{{\mathbb G}}
\def\HB{{\mathbb H}}
\def\IB{{\mathbb I}}
\def\JB{{\mathbb J}}
\def\KB{{\mathbb K}}
\def\LB{{\mathbb L}}
\def\MB{{\mathbb M}}
\def\NB{{\mathbb N}}
\def\OB{{\mathbb O}}
\def\PB{{\mathbb P}}
\def\QB{{\mathbb Q}}
\def\RB{{\mathbb R}}
\def\SB{{\mathbb S}}
\def\TB{{\mathbb T}}
\def\UB{{\mathbb U}}
\def\VB{{\mathbb V}}
\def\WB{{\mathbb W}}
\def\XB{{\mathbb X}}
\def\YB{{\mathbb Y}}
\def\ZB{{\mathbb Z}}
\def\wS{{\widetilde{\S}}}
\def\wSS{{\widetilde{S}}}
\def\wb{{\widetilde{\bbb}}}
\def\wbb{{\widetilde{b}}}
\def\wQ{{\widetilde{\Q}}}
\def\ob{{\overline{\bbb}}}
\def\obb{{\overline{b}}}
\def\oQ{{\overline{\Q}}}
\def\op{{\overline{\p}}}
\def\opp{{\overline{p}}}
\def\cW{{\mathcal{W}}}
\def\cP{{\mathcal{P}}}
\def\al{{\boldsymbol\alpha}}

\newcommand{\bigO}[1]{\ensuremath{\operatorname{\OM}(#1)}}
\newtheorem{remark}{Remark}

\begin{abstract}
Link prediction is a fundamental problem in many graph based applications, such as protein-protein interaction prediction. Graph neural network~(GNN) has recently been widely used for link prediction. However, existing GNN based link prediction~(\mbox{GNN-LP}) methods suffer from scalability problem during training for large-scale graphs, which has received little attention by researchers. In this paper, we first give computation complexity analysis of existing GNN-LP methods, which reveals that the scalability problem stems from their symmetric learning strategy adopting the same class of GNN models to learn representation for both head and tail nodes. Then we propose a novel method, called \underline{a}sym\underline{m}etric \underline{l}earning~(AML), for GNN-LP. The main idea of AML is to adopt a GNN model for learning head node representation while using a multi-layer perceptron~(MLP) model for learning tail node representation. Furthermore, AML proposes a row-wise sampling strategy to generate mini-batch for training, which is a necessary component to make the asymmetric learning strategy work for training speedup. To the best of our knowledge, AML is the first \mbox{GNN-LP} method adopting an asymmetric learning strategy for node representation learning. Experiments on three real large-scale datasets show that \mbox{AML} is $1.7{\times}{\sim}7.3{\times}$ faster in training than baselines with a symmetric learning strategy, while having almost no accuracy loss.
\end{abstract}

\section{Introduction}
Link prediction~\cite{DBLP:journals/jasis/Liben-NowellK07}, a fundamental problem in many graph based applications, aims to predict the existence of a link that has not been observed. Link prediction problem widely exists in real applications, like drug response prediction~\cite{DBLP:conf/bcb/StanfieldCK17}, protein-protein interaction prediction~\cite{qi2006evaluation}, friendship prediction in social networks~\cite{adamic2003friends}, knowledge graph completion~\cite{DBLP:journals/pieee/Nickel0TG16,DBLP:journals/tkdd/RossiBFMM21,DBLP:journals/corr/abs-2302-05044} and product recommendation in recommender systems~\cite{DBLP:journals/computer/KorenBV09}. Its increased importance in real applications also promotes a great interest in research for link prediction algorithms in the machine learning community.

Link prediction algorithms have been studied for a long time~\cite{lu2011link,VJF2017,DBLP:journals/inis/SamadQNIA20a}, while learning based algorithms are one dominant class in the past decades. The main idea of learning based algorithms is to learn a deterministic model~\cite{DBLP:journals/computer/KorenBV09,DBLP:conf/pkdd/MenonE11,zhang2017weisfeiler} or a probabilistic model~\cite{DBLP:conf/nips/SalakhutdinovM07,DBLP:journals/ftml/GoldenbergZFA09,guimera2009missing} to fit the observed data. In most learning based algorithms, models learn or generate a representation for each node~\cite{DBLP:journals/tkde/ZhangXKNZ21}, which is used to generate a score or probability of link existence. Traditional learning based algorithms typically do not adopt graph neural network~(GNN) for node representation learning. 
Although these non-GNN based learning algorithms have achieved much progress in many applications, they are less expressive than GNN in node representation learning.

\begin{figure}[t]
	\centering
	\includegraphics[width=0.70\linewidth]{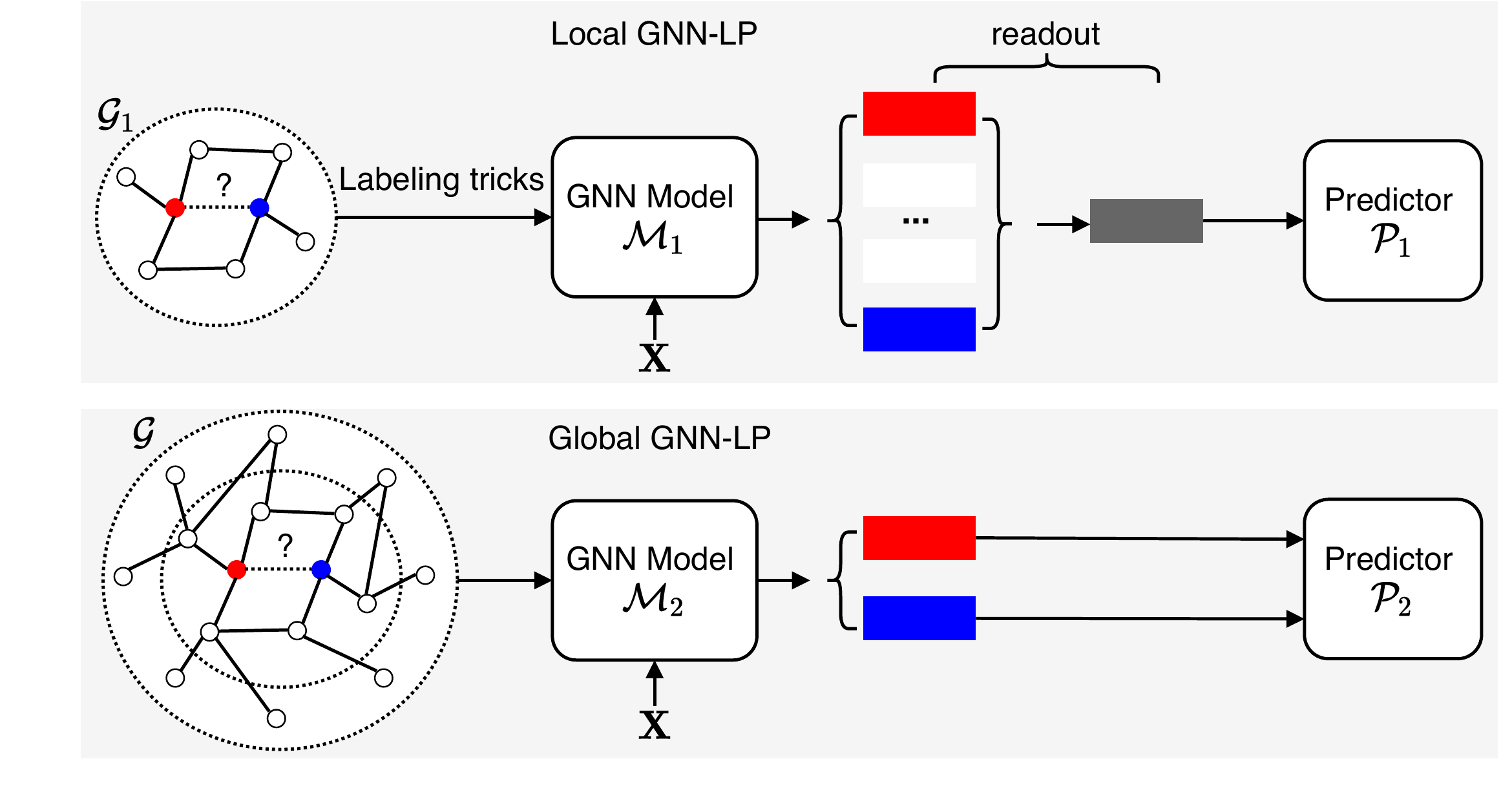}
	\caption{An illustration of existing representative GNN-LP methods. A circle denotes a node. $\mathcal{G}$ denotes the input graph. $\mathcal{G}_1$ denotes a subgraph extracted from $\mathcal{G}$. The dashed line between the red and blue circles denotes the target link we want to predict. $\X$ denotes the node feature matrix. The red and blue rectangles denote the representation of the red and blue circles, respectively. The gray rectangle denotes the subgraph representation of $\mathcal{G}_1$ generated by applying a readout function on representation of the nodes within the subgraph.}
	\label{fig:example}	
\end{figure}

Recently, graph neural network based link prediction~(GNN-LP) methods
have been proposed and become one of the most popular algorithms due to their superior performance in accuracy. The key to the success of GNN-LP methods is that they learn node representation from graph structure and node features in a unified way with GNN, which is a major difference between them and traditional non-GNN based learning algorithms. Existing \mbox{GNN-LP} methods mainly include local methods~\cite{DBLP:conf/nips/ZhangC18,DBLP:conf/nips/LiWWL20,DBLP:conf/nips/ZhangLXWJ21,DBLP:conf/aaai/YouGYL21} and global methods~\cite{DBLP:journals/corr/KipfW16a,DBLP:conf/ijcai/PanHLJYZ18,DBLP:conf/nips/HasanzadehHNDZQ19,DBLP:conf/nips/YunKLKK21}. Local methods apply GNN to subgraphs that capture local structural information. Specifically, they first extract an enclosed $k$-hop subgraph for each link and then use various labeling tricks~\cite{DBLP:conf/nips/ZhangLXWJ21} to capture the relative positions of nodes in the subgraph. After that, they learn node representation by applying a GNN model to the labeled subgraphs, and then they extract subgraph representation with a readout function~\cite{DBLP:conf/icml/GilmerSRVD17} for prediction. Global methods learn node representation by directly applying a GNN model to the global graph and then make prediction based on head and tail node representation. Although there exists difference in local methods and global methods, all existing GNN-LP methods have a common characteristic that they adopt a symmetric learning strategy for node representation learning. In particular, they adopt the same class of GNN models to learn representation for both head and tail nodes. An illustration of existing representative GNN-LP methods is presented in Figure~\ref{fig:example}. Although existing GNN-LP methods have made much progress in learning expressive models, they suffer from scalability problem during training for large-scale graphs, which has attracted little attention by researchers.

In this paper, we propose a novel method, called \underline{a}sym\underline{m}etric \underline{l}earning~(AML), for GNN-LP. The contributions of this paper are listed as follows:
\begin{itemize}
	\item We give computation complexity analysis of existing GNN-LP methods, which reveals that the scalability problem stems from their symmetric learning strategy adopting the same class of GNN models to learn representation for both head and tail nodes. 
	\item AML is the first GNN-LP method adopting an asymmetric learning strategy for node representation learning. 
	\item AML proposes a row-wise sampling strategy to generate mini-batch for training, which is a necessary component to make the asymmetric learning strategy work for training speedup.
	\item Experiments on three real large-scale datasets show that \mbox{AML} is $1.7{\times}{\sim}7.3{\times}$ faster in training than baselines with a symmetric learning strategy, while having almost no accuracy loss.
\end{itemize}

\section{Preliminary}\label{sec:preliminary}
In this section, we introduce notations and some related works for link prediction.

\paragraph{Notations.} We use a boldface uppercase letter, such as $\B$, to denote a matrix. We use a boldface lowercase letter, such as $\b$, to denote a vector. $\B_{i*}$ and $\B_{*j}$ denote the $i$th row and the $j$th column of $\B$, respectively. $\X{\in}\mathbb{R}^{N\times u}$ denotes the node feature matrix, where $u$ is the feature dimension and $N$ is the number of nodes. $\A{\in}\{0,1\}^{N\times N}$ denotes the adjacency matrix of a graph $\mathcal{G}$. $A_{ij}{=}1$ iff there is an edge from node $i$ to node $j$, otherwise $A_{ij}{=}0$. $L$ denotes the number of layers for GNN models. $\mathcal{E}$ denotes the set of links for training. For a link $(i,j)$, we call node $i$ a \emph{head node} and call node $j$ a \emph{tail node}. 

\paragraph{Graph Neural Network.} GNN~\cite{MGF2005,DBLP:journals/tnn/ScarselliGTHM09,DBLP:journals/corr/BrunaZSL13,DBLP:conf/iclr/KipfW17,DBLP:conf/nips/HamiltonYL17,GAT2018,DBLP:conf/kdd/JinL0AT22} is a class of models for learning over graph data. In GNN, nodes can iteratively encode their first-order and high-order neighbor information in the graph through message passing between neighbor nodes~\cite{DBLP:conf/icml/GilmerSRVD17}. Due to the iteratively dependent nature, the computation complexity for a node exponentially increases with iterations. Although some works~\cite{DBLP:conf/nips/ZouHWJSG19,DBLP:conf/iclr/ZengZSKP20,DBLP:conf/ijcai/YaoL21,DBLP:conf/kdd/GaoWJ18} propose solutions for the above problem of exponential complexity, the computation complexity of GNN is still much higher than that of a multi-layer perceptron~(MLP).

\paragraph{Graph Neural Network based Link Prediction.} Benefited from the powerful ability of GNN in modeling graph data, GNN-LP methods are more expressive than traditional non-GNN based learning algorithms in node representation learning. \mbox{GNN-LP} methods include two major classes, local methods
and global methods
. For local methods, different methods vary in the labeling tricks they use, which mainly include double radius node labeling~(DRNL)~\cite{DBLP:conf/nips/ZhangC18}, distance encoding~(DE)~\cite{DBLP:conf/nips/LiWWL20}, partial zero-one labeling trick~\cite{DBLP:conf/aaai/YouGYL21} and zero-one labeling trick~\cite{DBLP:conf/nips/ZhangLXWJ21}. As shown in~\cite{DBLP:conf/nips/ZhangLXWJ21}, local methods with DRNL and DE perform better than other methods. However, one bottleneck for DRNL and DE is that they need to compute the shortest path distance~(SPD) between target nodes and other nodes in subgraphs, and computing SPD is time-consuming during the training process~\cite{DBLP:conf/nips/ZhangLXWJ21}. Although we can compute SPD in the preprocessing step, costly storage overhead for subgraphs occurs instead. For global methods~\cite{DBLP:journals/corr/KipfW16a,DBLP:conf/ijcai/PanHLJYZ18,DBLP:conf/nips/HasanzadehHNDZQ19,DBLP:conf/nips/YunKLKK21}, they mainly apply different GNN models on the global graphs to generate node representation. Almost all existing GNN-LP methods, including both local methods and global methods, adopt a symmetric learning strategy which utilizes the same class of GNN models to learn representation for both head and tail nodes. 

\paragraph{Non-GNN based Methods.} Besides GNN-LP methods, WLNM~\cite{zhang2017weisfeiler} and SUREL \cite{DBLP:journals/corr/abs-2202-13538} also show competitive performance in accuracy. The main difference between them and GNN-LP methods is that they do not apply GNN to learn from graphs or subgraphs. For example, WLNM applies an MLP model to learn subgraph representation from the adjacency matrices of extracted subgraphs. SUREL proposes an alternative sampler for subgraph extraction and applies a sequential model, like recurrent neural networks~(RNNs), to learn subgraph representation.

In general, GNN-LP methods have higher accuracy than non-GNN based methods~\cite{DBLP:conf/nips/ZhangLXWJ21}, but suffer from scalability problem during training for large-scale graphs. Although global \mbox{GNN-LP} methods are more efficient than local GNN-LP methods, they still have high computation complexity in training due to the adopted symmetric learning strategy. This motivates our work in this paper.

\section{Asymmetric Learning for GNN-LP}

Like most deep learning methods, GNN-LP methods are typically trained in a mini-batch manner. Suppose the number of links in the training set is $|\mathcal{E}|$. Then existing \mbox{GNN-LP} methods with symmetric learning need to perform $2|\mathcal{E}|$ times of GNN computation within each epoch. In particular, $|\mathcal{E}|$ times of GNN computation are for head nodes and another $|\mathcal{E}|$ times of GNN computation are for tail nodes. It is easy to verify that $|\mathcal{E}|$ times of GNN computation are inevitable for both head and tail node representation learning with a symmetric learning strategy. Since GNN is of exponential computation complexity and $|\mathcal{E}|$ is of a considerably large value, existing GNN-LP methods incur a huge computation burden for large-scale graphs. 



\begin{figure}[t]
	\centering
	\includegraphics[width=0.6\linewidth]{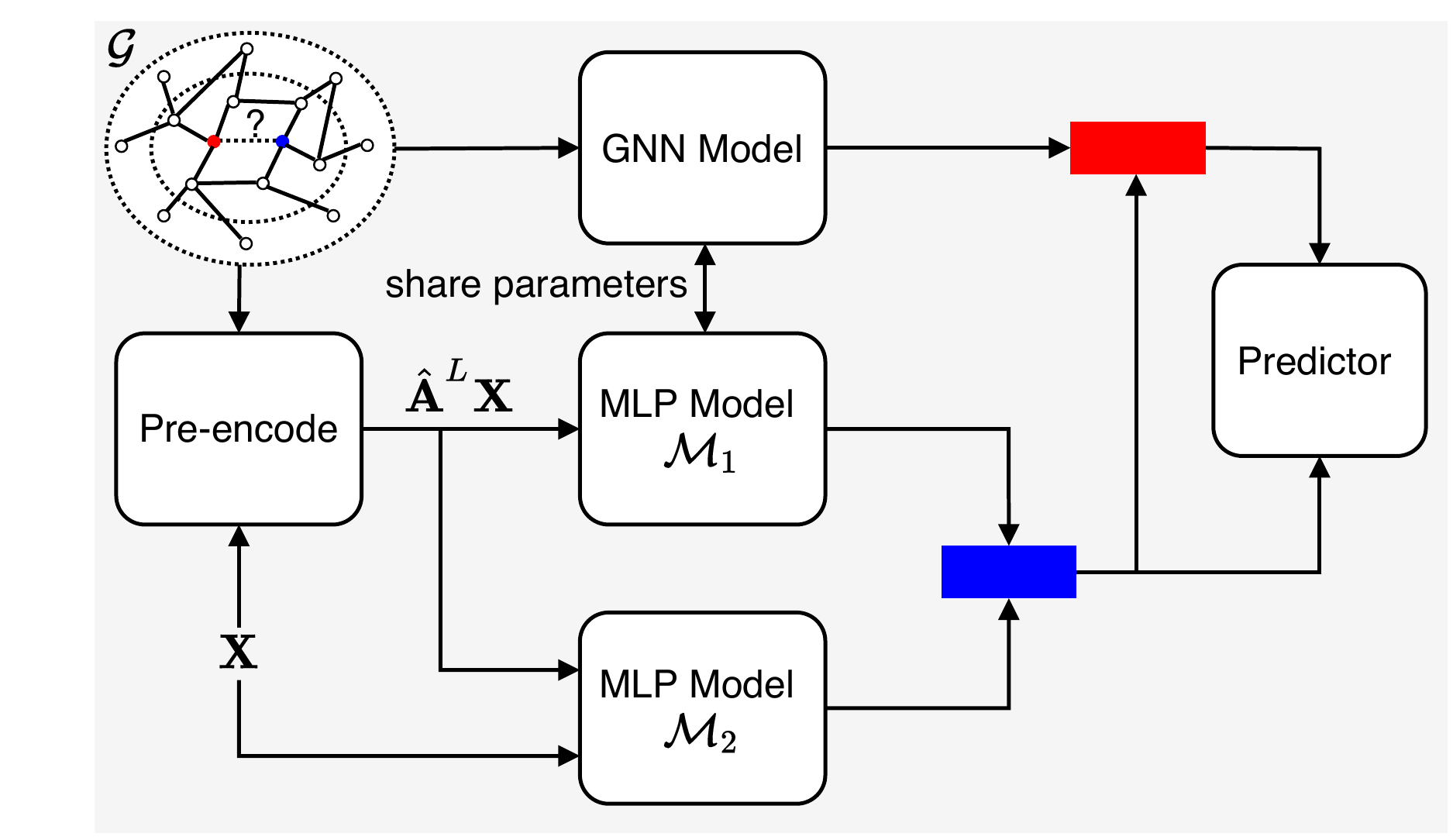}
	\caption{An illustration of AML. The MLP model $\mathcal{M}_1$ is for transfering knowledge from head nodes to tail nodes by sharing parameters with the GNN model. The MLP model $\mathcal{M}_2$ is for learning over the residual between $\A^L\X$ and $\X$. AML obtains the tail node representation, marked with the blue rectangle, by summing up the outcomes of $\mathcal{M}_1$ and $\mathcal{M}_2$. AML obtains the head node representation, marked with the red rectangle, by summing up the tail node representation and the outcomes of the GNN model. The predictor is for generating predictions according to the input vector representation.}
	\label{fig:arch}	
\end{figure}

To solve the scalability problem caused by symmetric learning, we propose AML which is illustrated in Figure~\ref{fig:arch}. The main idea of AML is to adopt a GNN model for learning head node representation while using a multi-layer perceptron~(MLP) model for learning tail node representation. Meanwhile, AML pre-encodes graph structure to avoid information loss for MLP. The following content of this section will present the details of AML.



\subsection{Node Representation Learning with AML}
We use $\U{\in}\mathbb{R}^{N\times r}$ to denote the representation of all head nodes and use $\V{\in}\mathbb{R}^{N\times r}$ to denote the representation of all tail nodes, where each row of $\U$ and $\V$ corresponds to a node. We apply a GNN model to learn representation for head nodes while using an MLP model to learn representation for tail nodes\footnote{We can also apply a GNN model to learn representation for tail nodes while using an MLP model to learn representation for head nodes. For convenience, we only present the technical details of one case. The technical details for the reversed case are the same as the presented one.}. 

We take SAGE~\cite{DBLP:conf/nips/HamiltonYL17}, one of the most representative GNN models, as an example to describe the details. Note that other GNN models can also be used in AML. Let $\hat{\A}$ denote the normalization of $\A$, which can be row-normalization, column-normalization or symmetric normalization. Formulas for one layer in SAGE are as follows:
\begin{align}
	&\U^{(\ell)}_{i*} = f\left(\hat{\A}_{i*}\U^{(\ell-1)}\W^{(\ell)}_1 + \U^{(\ell-1)}_{i*}\W^{(\ell)}_2\right),\label{eq:head_gnn1}
\end{align}
where $\ell$ is layer number, $f(\cdot)$ is an activation function, $\W^{(\ell)}_1{\in}\mathbb{R}^{r\times r}$ and $\W^{(\ell)}_2{\in}\mathbb{R}^{r\times r}$ are learnable parameters at layer $\ell$. $\U^{(\ell)}{\in}\mathbb{R}^{N\times r}$ is the node representation at layer $\ell$ and $\U^{(0)} {=} \X$. In~(\ref{eq:head_gnn1}), node $i$ encodes neighbor information via $\hat{\A}_{i*}\U^{(\ell-1)}\W^{(\ell)}_1$ and updates its own message together with $\U^{(\ell-1)}_{i*}\W^{(\ell)}_2$. 

Different from existing GNN-LP methods which adopt the same class of GNN models to learn representation for both head and tail nodes, AML applies an MLP model to learn representation for tail nodes. However, naively applying MLP for tail nodes will deteriorate the accuracy. Hence, as in~\cite{DBLP:conf/nips/KlicperaWG19,DBLP:conf/icml/WuSZFYW19,DBLP:conf/nips/ChenWDL00W20}, we first pre-encode graph structure information into node features in the preprocessing step. The pre-encoding step is as follows:
\begin{align}
	\tilde{\V}^{(0)} = \hat{\A}^L\X.\label{eq:prepro}
\end{align}
To further improve the representation learning for tail nodes, we propose to transfer knowledge from head nodes to tail nodes by sharing parameters. The formula is as follows:
\begin{align}
	\tilde{\V}^{(\ell)} = f\left(\tilde{\V}^{(\ell-1)}\W_1^{(\ell)} + \tilde{\V}^{(\ell-1)}\W_2^{(\ell)}\right),\label{eq:tail_mlp1}
\end{align}
where we perform knowledge transfer between head nodes and tail nodes by sharing $\W_1^{(\ell)}$ and $\W_2^{(\ell)}$. Since sharing parameters somewhat restricts the expressiveness of $\tilde{\V}^{(L)}$, we propose to apply an MLP model to learn over the residual $\Delta^{(0)}$. Adding the residual $\Delta^{(L)}$ to $\tilde{\V}^{(L)}$, we obtain the representation for tail nodes which is shown as follows:
\begin{align}
	&\Delta^{(0)} = \X - \hat{\A}^L\X,\label{eq:tail_mlp4}\\
	&\Delta^{(\ell)} = f\left(\Delta^{(\ell-1)}\W^{(\ell)}\right),\label{eq:tail_mlp2}\\
	&\V^{(L)} = \tilde{\V}^{(L)} + \Delta^{(L)},\label{eq:tail_mlp3}
\end{align}
where $\W^{(\ell)}\in\mathbb{R}^{r\times r}$ is a learnable parameter at layer $\ell$.

Note that if BatchNorm~\cite{DBLP:conf/icml/IoffeS15} is applied to $\U^{(\ell)}$ and $\tilde{\V}^{(\ell)}$, we keep individual parameters of BatchNorm for $\U^{(\ell)}$ and $\tilde{\V}^{(\ell)}$. Since $\U^{(\ell)}$ and $\tilde{\V}^{(\ell)}$ have different scales and lie in different representation spaces, it is reasonable to keep individual parameters of BatchNorm for them. Here pre-encoding step only needs to be performed once and the resulting $\hat{\A}^L\X$ in~(\ref{eq:prepro}) can be saved for the entire training process. 


\subsection{Learning Objective of AML}

According to the definitions in~(\ref{eq:head_gnn1}) and~(\ref{eq:tail_mlp3}), modeling links with $\U^{(L)}$ and $\V^{(L)}$ can capture directed relations while being a little difficult to capture undirected relations. Our solutions are twofold. Firstly, we formulate each undirected link by two directed ones with opposite directions. Secondly, motivated by the work in~\cite{DBLP:conf/ijcai/LiYZ11}, in AML we propose to model both \emph{homophily} and \emph{stochastic equivalence}~\cite{DBLP:conf/nips/Hoff07}. As a result, the formulas for the prediction of a pair $(i,j)$ are as follows:
\begin{align}
	&\U = \U^{(L)} + \V^{(L)},\quad \V = \V^{(L)},\label{eq:uv}\\
	&S_{ij} = f_{\bm{\Theta}}\left(\U_{i*} \odot \V_{j*}\right)\label{eq:pred},
\end{align}
where $\U$ and $\V$ are representation for head and tail nodes, respectively. $\V^{(L)}$ in $\U$ is included to model the \emph{homophily} feature in graph data. $S_{ij}$ is the prediction for the pair $(i,j)$. $\odot$ denotes element-wise multiplication. $f_{\bm{\Theta}}(\cdot)$ is an MLP model with parameter $\bm{\Theta}$. 

Given node representation $\U$ and $\V$, the learning objective of link prediction is as follows:
\begin{align}
	\min_{\mathcal{W}} \frac{1}{|\mathcal{E}|}\sum_{(i,j)\in\mathcal{E}} f_{loss}(S_{ij}, Y_{ij}) + \frac{\lambda}{2} \sum_{\W\in\mathcal{W}}\lVert \W\rVert_F^2,\label{obj:full}
\end{align}
where $\mathcal{E}$ denotes the training set. $\mathcal{W}$ is the set of all learnable parameters. $Y_{ij}$ is the ground-truth label for the pair $(i,j)$. $f_{loss}(\cdot, \cdot)$ is a loss function, such as cross-entropy loss. $\lambda$ is a coefficient for the regularization term of $\mathcal{W}$. $\lVert\cdot\rVert_F$ denotes the Frobenius norm of a matrix.



\subsection{Row-wise Sampling for Generating Mini-Batch}



Like most deep learning methods, existing GNN-LP methods are typically trained in a mini-batch manner. But existing GNN-LP methods adopt an edge-wise sampling strategy to generate mini-batch for training. In particular, they first randomly sample a mini-batch of edges $\mathcal{E}_\Omega$ from $\mathcal{E}$ at each iteration and then optimize the objective function based on $\mathcal{E}_\Omega$. For example, if we adopt the edge-wise sampling strategy for the objective function of AML in (\ref{obj:full}), the corresponding learning objective at each iteration will be as follows:
\begin{align}
	\min_{\mathcal{W}} \frac{1}{|\mathcal{E}_{\Omega}|}\sum_{(i,j)\in\mathcal{E}_{\Omega}} f_{loss}(S_{ij}, Y_{ij}) + \frac{\lambda}{2} \sum_{\W\in\mathcal{W}}\lVert \W\rVert_F^2.\label{obj:random}
\end{align}
Suppose $\mathcal{E}_\Omega{=}\{(i_1, j_1), {\cdots}, (i_B, j_B)\}$ with $B$ as the mini-batch size. We respectively use $C$ and $M$ to denote the computation complexity for generating a node representation by GNN and MLP. Since $\mathcal{E}_\Omega$ is edge-wise randomly sampled from $\mathcal{E}$ and $N$ is of a large value for large-scale graphs, $(i_1, {\cdots}, i_B)$ will have a relative small number of repeated nodes. Then the edge-wise sampling strategy for AML has a computation complexity of $\mathcal{O}(|\mathcal{E}|{\cdot}(C{+}M))$ for each epoch, which has the same order of magnitude in computation complexity as existing GNN-LP methods and hence is undesirable.

To solve this high computation complexity problem of edge-wise sampling strategy adopted by existing GNN-LP methods, in AML we propose a row-wise sampling strategy for generating mini-batch. More specifically, we first sample a number of row indices $\mathcal{V}_\Omega$ (head nodes) from $\{1,2,\cdots,N\}$ for each mini-batch iteration. Then we construct the mini-batch $\mathcal{E}_\Omega$ as follows: 
\begin{align}
	\mathcal{E}_\Omega=\{(i,j)|i\in\mathcal{V}_\Omega\land(i,j)\in\mathcal{E}\}.\label{eq:rws}
\end{align}
By using this row-wise sampling strategy, AML has a computation complexity of $\mathcal{O}(|\mathcal{V}_\Omega|{\cdot}C+ (|\mathcal{V}_\Omega|/N)|\mathcal{E}|\cdot M)$ for each mini-batch iteration. Since $\mathcal{V}_\Omega$ iterates over $\{1,{\cdots},N\}$ for $N/|\mathcal{V}_\Omega|$ times to go through the whole training set, this row-wise sampling strategy for AML has a computation complexity of $\mathcal{O}(N{\cdot}C{+} |\mathcal{E}|{\cdot}M)$ for each epoch. Therefore, the row-wise sampling strategy enables AML to decouple the factor of $|\mathcal{E}|$ from the computation complexity of GNN, leading to a complexity reduction by orders of magnitude compared to the edge-wise sampling strategy.

Algorithm 1 summarizes the whole learning algorithm for AML.


\begin{algorithm}[!t]
	\caption{Learning Algorithm for AML}
	\label{algo:rws-AML}	
	\begin{algorithmic}[1]
		\REQUIRE $N$ (number of nodes in the input graph), $L$ (number of model layers), $\mathcal{E}$ (the training set), $B$ (mini-batch size), $T$ (maximum number of epoches).
		\ENSURE $\mathcal{W}$ (model parameters).
		\STATE Pre-encoding graph structure by~(\ref{eq:prepro});
		\STATE $\tilde{B}=B / (|\mathcal{E}|/N)$;
		\FOR {$t=1:T$}
		\FOR {$q=1:(N/\tilde{B})$}
		\STATE Sample $\mathcal{V}_\Omega$ of $\tilde{B}$ head nodes from $\{1,{\cdots},N\}$;
		\STATE Generate $\mathcal{E}_\Omega$ with $\mathcal{V}_\Omega$ by~(\ref{eq:rws});
		\STATE Compute $\U_{i*}^{(L)}$ for each node $i$ in $\mathcal{V}_\Omega$ by~(\ref{eq:head_gnn1});
		\STATE Compute $\V_{j*}^{(L)}$ for each node $j$ in $\mathcal{E}_\Omega$ by~(\ref{eq:tail_mlp1})-(\ref{eq:tail_mlp3});
		\STATE Compute $\U_{i*}$ and $\V_{j*}$ for each $(i,j)$ in $\mathcal{E}_\Omega$ by~(\ref{eq:uv});
		\STATE Compute $S_{ij}$ for each $(i,j)$ in $\mathcal{E}_\Omega$ by~(\ref{eq:pred});
		\STATE Update model parameters $\mathcal{W}$ by optimizing~(\ref{obj:random});  \textcolor{gray}{/*Note that the objective function of row-wise sampling is the same as that of edge-wise sampling in~(\ref{obj:random}).*/}
		\ENDFOR
		\ENDFOR
	\end{algorithmic}
\end{algorithm}

\begin{table*}[t]
	\centering
	\caption{Complexity analysis. $L$ is the number of model layers. $s{=}\lVert\A\rVert_0/N$ is the average number of neighbors for a node in $\mathcal{G}$. $r$ is the dimension of node representation. $|\mathcal{E}|$ is the number of links in the training set $\mathcal{E}$. $N$ is the number of nodes in graph $\mathcal{G}$. $k$ in local GNN-LP is the number of hops for the enclosed subgraphs. ``RWS'' denotes row-wise sampling. ``w/ RWS'' in local GNN-LP and global GNN-LP means generating mini-batch with RWS. ``w/o RWS'' in AML means generating mini-batch without RWS but with an edge-wise sampling strategy.}
	\label{tb:analysis}
	\begin{tabular}{lccccc}
		\toprule
		Method&Computation complexity\\
		\midrule
		Local GNN-LP &$\mathcal{O}\left(2Ls^{k}r^2{\cdot} |\mathcal{E}|\right)$\\
		Local GNN-LP (w/ RWS) &$\mathcal{O}\left(2Ls^{k}r^2{\cdot} |\mathcal{E}|\right)$\\
		Global GNN-LP &$\mathcal{O}\left(2s^{L}r^2{\cdot} |\mathcal{E}|\right)$\\
		Global GNN-LP (w/ RWS) &$\mathcal{O}\left(s^{L}r^2{\cdot} \left(|\mathcal{E}| + N\right)\right)$\\
		\midrule
		AML (w/o RWS) &$\mathcal{O}\left(\left(s^{L}r^2 + Lr^2\right){\cdot} |\mathcal{E}|\right)$\\
		AML &$\mathcal{O}\left(s^{L}r^2{\cdot} N + Lr^2{\cdot} |\mathcal{E}|\right)$\\
		\bottomrule
	\end{tabular}
\end{table*}

\subsection{Complexity Analysis}
The computation complexity for different methods are summarized in Table~\ref{tb:analysis}.
For large-scale graphs, we often have $s^L{<}N$ and $s^k{<}N$. Typically, $s^Lr^2$ has the same order of magnitude as $Ls^kr^2$. According to~(\ref{eq:head_gnn1}), GNN has a computation complexity of $C{=}\mathcal{O}\left(s^L{\cdot} r^2\right)$ to generate a node representation while the corresponding computation complexity of MLP is $M{=}\mathcal{O}\left(L{\cdot} r^2\right)$ according to~(\ref{eq:tail_mlp1})-(\ref{eq:tail_mlp3}). It is easy to verify that $M{\ll}C$. Although many works~\cite{DBLP:conf/nips/HamiltonYL17,DBLP:conf/nips/ZouHWJSG19,DBLP:conf/iclr/ZengZSKP20,DBLP:conf/ijcai/YaoL21} have proposed solutions to reduce $C$, $C$ is still much larger than $M$. Here we suppose all methods are trained in a mini-batch manner which has been adopted by almost all deep learning models including GNN.

From Table~\ref{tb:analysis}, we can get the following results. Firstly, AML has a computation complexity of $\mathcal{O}(N{\cdot}C{+} |\mathcal{E}|{\cdot}M)$, which is much lower than $\mathcal{O}(2|\mathcal{E}|{\cdot}C)$ of existing GNN-LP methods. Secondly, even with our proposed row-wise sampling strategy, existing GNN-LP methods still have a computation complexity of $\mathcal{O}((|\mathcal{E}|{+}N){\cdot}C)$, without change in the order of magnitude. The reason is that they still need to perform $|\mathcal{E}|$ times of GNN computation for tail nodes within each epoch.

\section{Experiments}
In this section, we evaluate AML and baselines on three real datasets. All methods are implemented with Pytorch~\cite{DBLP:conf/nips/PaszkeGMLBCKLGA19} and Pytorch-Geometric Library~\cite{Fey/Lenssen/2019}. All experiments are run on an NVIDIA RTX A6000 GPU server with 48 GB of graphics memory.



\begin{table}[t]
	\small
	\centering
	\caption{Statistics of datasets.}
	\label{tb:statistics}
	\begin{tabular}{lccccccccc}
		\toprule
		Datasets & ogbl-collab & ogbl-ppa & ogbl-citation2\\
		\midrule
		\#Nodes  & 235,868 & 576,289 & 2,927,963 \\
		\#Edges	 & 1,285,465 & 30,326,273 & 30,561,187\\
		Features/Node & 128 & 128 & 128\\
		\#Training links & 1,179,052 & 21,231,931 & 30,387,995\\
		\#Validation links & 160,084 & 9,062,562 & 86,682,596\\
		\#Test links & 146,329 & 6,031,780 & 86,682,596\\
		Metric & Hits@50 & Hits@100 & MRR\\
		\bottomrule
	\end{tabular}
\end{table}

\subsection{Settings}
\paragraph{Datasets} Datasets for evaluation include ogbl-collab\footnote{In ogbl-collab, there are data leakage issues in the provided graph adjacency matrix $\A$. We remove those positive links in the validation and testing set from $\A$.}, ogbl-ppa and ogbl-citation2\footnote{https://ogb.stanford.edu/docs/linkprop/}. The first two are medium-scale datasets with hundreds of thousands of nodes. The last one is a large-scale dataset with millions of nodes. For ogbl-ppa, since the provided node features are uninformative, we apply matrix factorization~\cite{DBLP:conf/pkdd/MenonE11} to generate new features for nodes. The first two datasets are for undirected link prediction, while the last one is for directed link prediction. The statistics of datasets are summarized in Table~\ref{tb:statistics}. Since most GNN-LP methods adopt the evaluation metrics provided by~\cite{DBLP:conf/nips/HuFZDRLCL20}, we also follow these evaluation settings.

\paragraph{Baselines} AML is actually a global GNN-LP method. We first compare AML with existing global GNN-LP baselines by adopting the same GNN for both AML and baselines. Since almost all existing global \mbox{GNN-LP} methods are developed based on the graph autoencoder framework proposed in~\cite{DBLP:journals/corr/KipfW16a}, we mainly adopt the GNNs under the graph autoencoder framework. In particular, we adopt SAGE~\cite{DBLP:conf/nips/HamiltonYL17} and GAT~\cite{GAT2018} as the GNNs for both AML and baselines, because SAGE and GAT are respectively representative non-attention based and attention based GNN models under the graph autoencoder framework. 

Then we compare AML with non-GNN baselines and local GNN-LP baselines. Non-GNN baselines include common neighbors~(CN)~\cite{newman2001clustering}, Adamic-Adar~(AA)~\cite{adamic2003friends}, Node2vec~\cite{grover2016node2vec} and matrix factorization~(MF)~\cite{DBLP:conf/pkdd/MenonE11}. Local GNN-LP baselines inlcude DE-GNN~\cite{DBLP:conf/nips/LiWWL20} and SEAL~\cite{DBLP:conf/nips/ZhangC18,DBLP:conf/nips/ZhangLXWJ21}. For local GNN-LP methods, we extract enclosed subgraphs in an online way to simulate large-scale settings by following the original work~\cite{DBLP:conf/nips/ZhangLXWJ21}.



\paragraph{Hyper-parameter Settings} Hyper-parameters include $L$~(layer number), $r$~(hidden dimension), $\lambda$~(coefficient for the regularization of parameters), $T$~(maximum number of epoches), $\eta$~(learning rate) and $B$~(mini-batch size). On ogbl-collab, $L{=}3$, $r{=}256$, $\lambda{=}0$, $T{=}400$, $\eta{=}0.001$ and $B{=}65,536$. On ogbl-ppa, $L{=}3$, $r{=}256$, $\lambda{=}0$, $T{=}50$, $\eta{=}0.01$ and $B{=}65,536$. On ogbl-citation2, $L{=}3$, $r{=}256$, $\lambda{=}0$, $T{=}50$, $\eta{=}0.005$ and $B{=}65,536$. We use Adam~\cite{DBLP:journals/corr/KingmaB14} as the optimizer. We use GraphNorm~\cite{DBLP:conf/icml/CaiLXHL021} to accelerate the training. We adopt BNS~\cite{DBLP:conf/ijcai/YaoL21} as the neighbor sampling strategy for large-scale training. In BNS, there are three hyper-parameters, including $\tilde{s}_0$, $\tilde{s}_1$ and $\delta$. $\tilde{s}_0$ denotes the number of sampled neighbors for nodes at output layer. $\tilde{s}_1$ denotes the number of sampled neighbors for nodes at other layers. $\delta$ denotes the ratio of blocked neighbors. On ogbl-collab, $\tilde{s}_0{=}7$, $\tilde{s}_1{=}2$, $\delta{=}1/2$. On ogbl-ppa, $\tilde{s}_0$ equals the number of all neighbors, $\tilde{s}_1{=}4$, $\delta{=}5/6$. On ogbl-citation2, $\tilde{s}_0{=}7$, $\tilde{s}_1{=}2$, $\delta{=}1/2$. We run each setting for 5 times and report the mean with standard deviation.

\begin{table*}[!t]
	\centering
	\small
	\caption{Comparison with global GNN-LP baselines. ``Time'' denotes the whole time to complete training. ``Gap'' denotes the accuracy of AML minus that of  baselines.}
	\label{tb:main-result}
	\begin{subtable}{\linewidth}
		\centering
		\caption{SAGE as the GNN model.}
		\label{tb:main-result-sage}
		\begin{tabular}{lccccccccccccc}
			\toprule
			\multirow{2}*{Methods} &\multicolumn{3}{c}{ogbl-collab} &\multicolumn{3}{c}{ogbl-ppa}\\
			\cmidrule{2-7}
			& Hits@50 (\%) $\uparrow$ & Gap  & Time (s) $\downarrow$ & Hits@100 (\%) $\uparrow$ & Gap & Time (s) $\downarrow$\\
			\midrule
			SAGE & 54.57 $\pm$ 0.82 & - & $7.9\times10^3$ &\textbf{50.13 $\pm$ 0.55} & - & $1.0\times10^5$\\
			AML & \textbf{57.26 $\pm$ 1.25} & $+2.69$ & $\bm{4.4\times10^3}$ & 49.73 $\pm$ 0.89 & $-0.40$ & $\bm{3.0\times10^4}$\\ 
			\bottomrule
		\end{tabular}
		\vskip 0.05in
		\begin{tabular}{lccccccccccccc}
			\toprule
			\multirow{2}*{Methods} &\multicolumn{3}{c}{ogbl-citation2}\\
			\cmidrule{2-4}
			& MRR (\%) $\uparrow$ & Gap & Time (s) $\downarrow$\\
			\midrule
			SAGE & 86.39 $\pm$ 0.15 & - & $7.3\times10^4$\\
			AML & \textbf{86.55 $\pm$ 0.06} & $+0.16$ & $\bm{1.0\times10^4}$\\ 
			\bottomrule
		\end{tabular}
	\end{subtable}
	\vskip 0.1in
	\begin{subtable}{\linewidth}
		\centering
		\caption{GAT as the GNN model.}
		\label{tb:main-result-gat}
		\begin{tabular}{lccccccccccc}
			\toprule
			\multirow{2}*{Methods} &\multicolumn{3}{c}{ogbl-collab} &\multicolumn{3}{c}{ogbl-ppa}\\
			\cmidrule{2-7}
			& Hits@50 (\%) $\uparrow$ & Gap & Time (s) $\downarrow$ & Hits@100 (\%) $\uparrow$ & Gap & Time (s) $\downarrow$ \\
			\midrule
			GAT & 56.43 $\pm$ 0.86 & - & $7.8\times10^3$ &49.71 $\pm$ 0.48 & - & $1.0\times10^5$ \\
			AML & \textbf{57.60 $\pm$ 0.71} & $+1.17$ & $\bm{4.5\times10^3}$ & \textbf{50.23 $\pm$ 0.78} & $+0.52$ & $\bm{3.2\times10^4}$ \\  
			\bottomrule
		\end{tabular}
		\vskip 0.05in
		\begin{tabular}{lccccccccccc}
			\toprule
			\multirow{2}*{Methods} &\multicolumn{3}{c}{ogbl-citation2}\\
			\cmidrule{2-4}
			& MRR (\%) $\uparrow$ & Gap & Time (s) $\downarrow$\\
			\midrule
			GAT & 86.50 $\pm$ 0.20 & - & $7.4\times10^4$\\
			AML & \textbf{86.70 $\pm$ 0.05} & $+0.20$ & $\bm{1.0\times10^4}$\\  
			\bottomrule
		\end{tabular}
	\end{subtable}
\end{table*}

\subsection{Comparison with Global GNN-LP Baselines}
\begin{figure*}[!t]
	\centering
	\begin{subfigure}[h]{\linewidth}
		\label{fig:result-sage}
		\centering
		\includegraphics[width=0.32\linewidth]{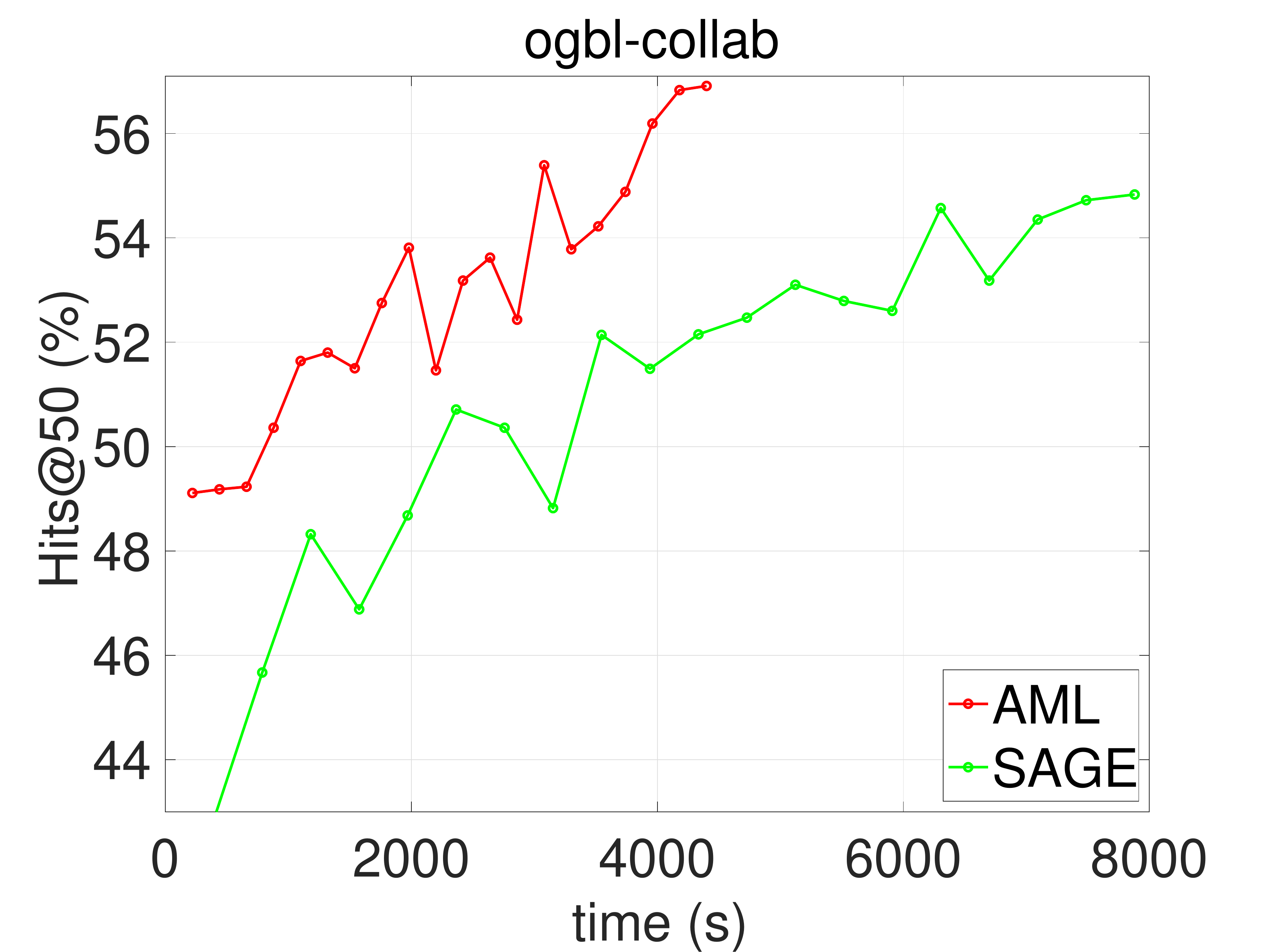}
		\hfill
		\includegraphics[width=0.32\linewidth]{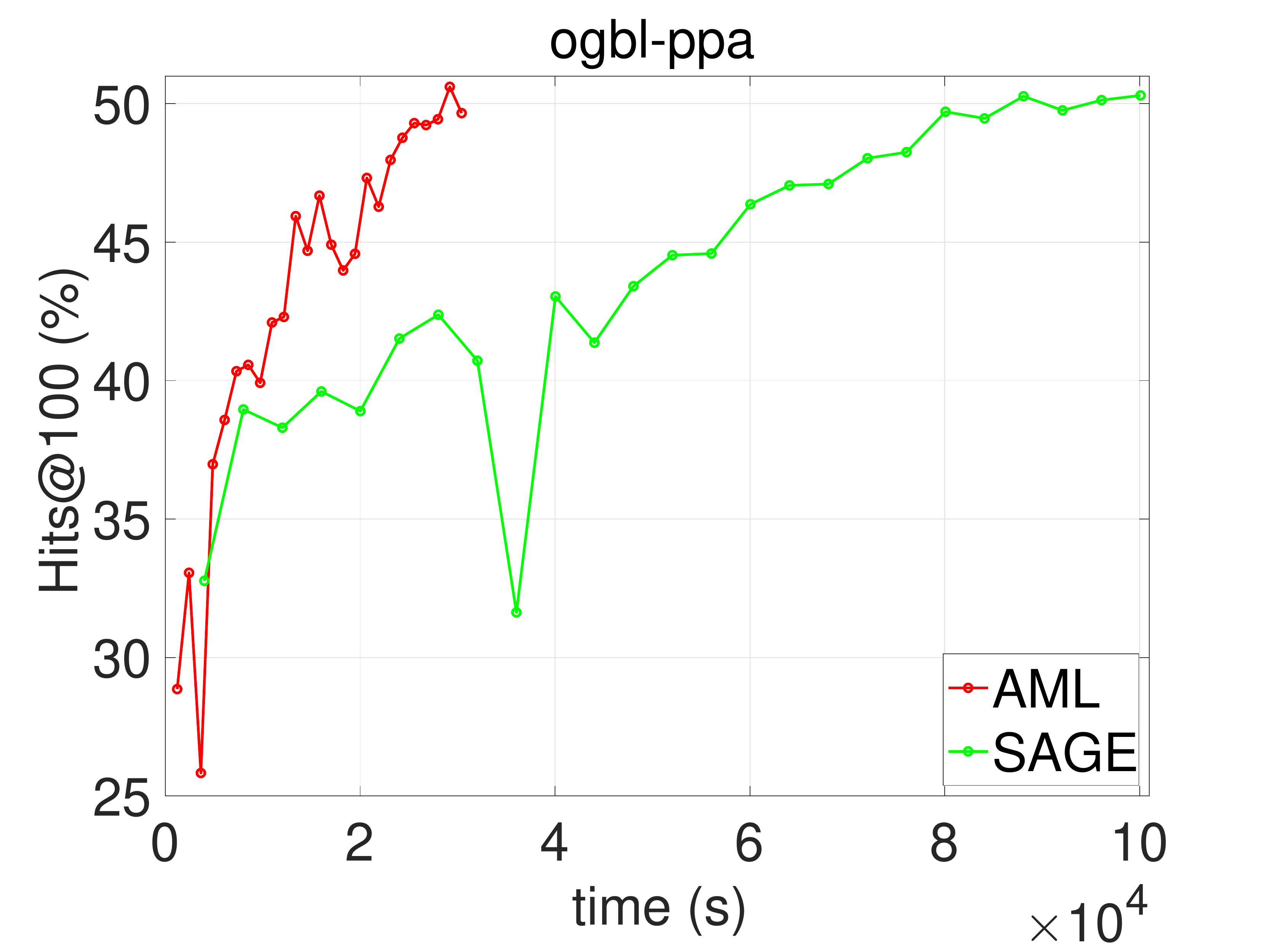}
		\hfill
		\includegraphics[width=0.32\linewidth]{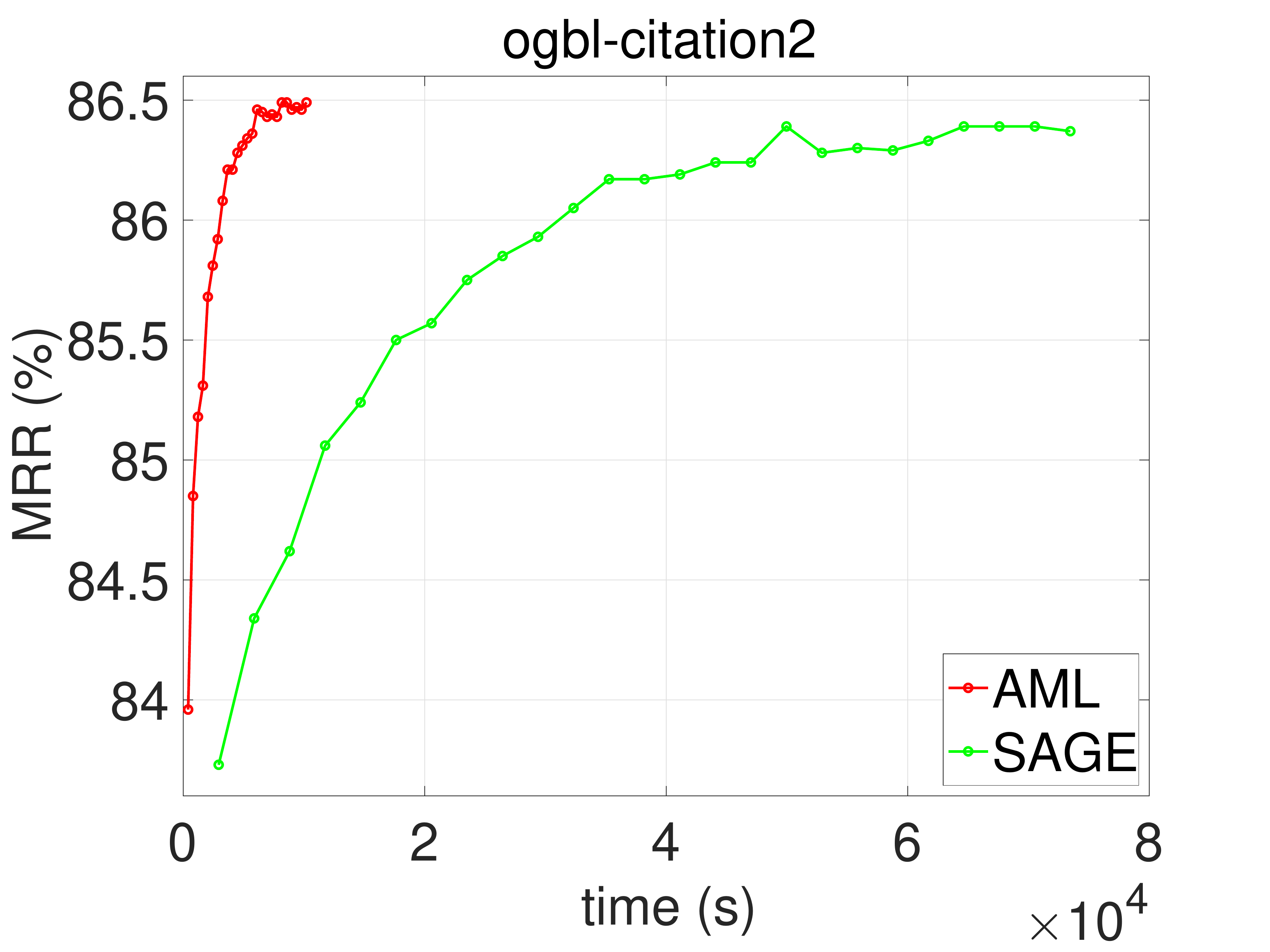}
		\caption{SAGE as the GNN model.}
	\end{subfigure}
	\vskip 0.1in
	\begin{subfigure}[h]{\linewidth}
		\label{fig:result-gat}
		\centering
		\includegraphics[width=0.32\linewidth]{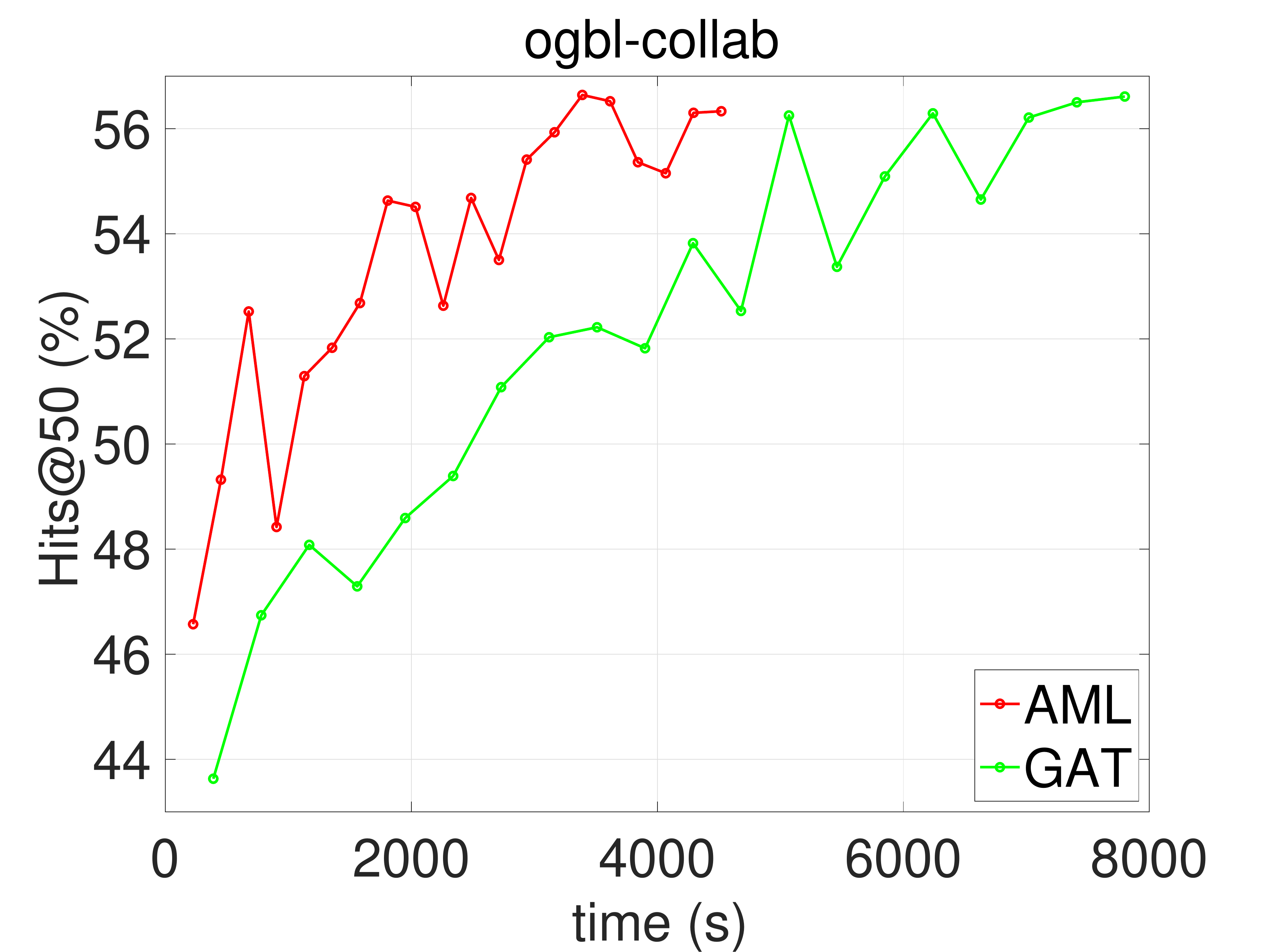}
		\hfill
		\includegraphics[width=0.32\linewidth]{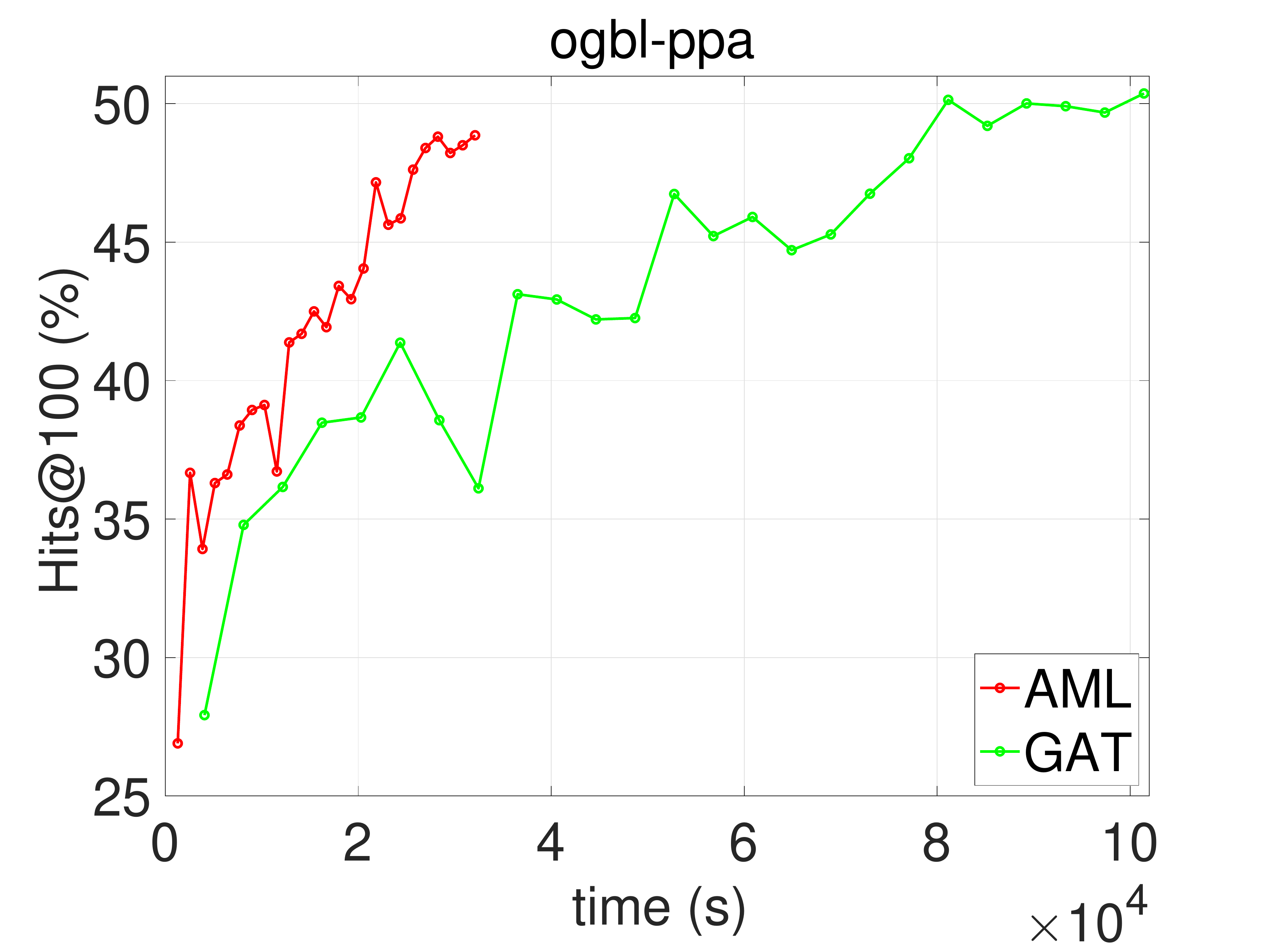}
		\hfill
		\includegraphics[width=0.32\linewidth]{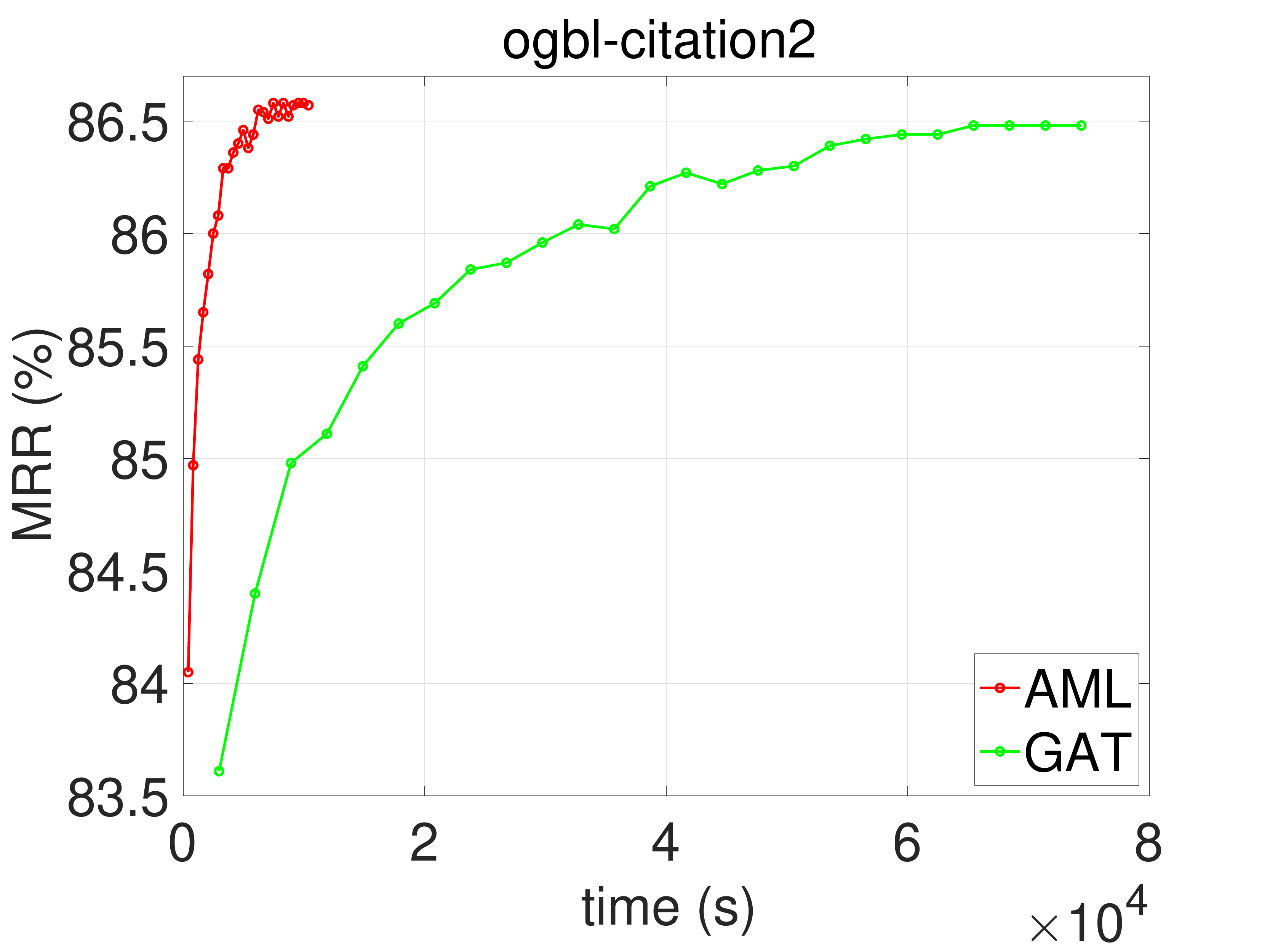}
		\caption{GAT as the GNN model.}
	\end{subfigure}
	\caption{Test accuracy-time curves of AML and global GNN-LP baselines.}
	\label{fig:result}
\end{figure*}

Comparison with global GNN-LP baselines is shown in Table~\ref{tb:main-result} and Figure~\ref{fig:result}. According to the results, we can draw the following conclusions. Firstly, AML has almost no accuracy loss in all cases compared to baselines\footnote{When AML's accuracy is within the standard deviation of baselines' accuracy, we say that AML achieves almost no accuracy loss compared to baselines.}.  For example, AML's accuracy is within the standard deviation of baselines' accuracy on ogbl-ppa and ogbl-citation2. Instead, AML achieves an accuracy gain of $1.17\%{\sim}2.69\%$ on ogbl-collab compared to baselines. Secondly, AML is about $1.7{\times}{\sim}7.3{\times}$ faster in training than baselines when achieving almost no accuracy loss. For example, \mbox{AML} is $1.7{\times}$ faster than baselines on ogbl-collab, $3.1{\times}$ faster on ogbl-ppa, and $7.3{\times}$ faster on ogbl-citation2. In particular, baselines need about 5.8 days and 4.2 days to get the mean of results while \mbox{AML} only needs 1.7 days and 0.6 days to get the mean of results, on ogbl-ppa and ogbl-citation2 respectively. Thirdly, the speedup of AML relative to baselines increases with the size of graphs. For example, the number of nodes in ogbl-collab, ogbl-ppa and ogbl-citation2 increases in an ascending order, and the speedup of AML relative to baselines increases in a consistent order on these three graph datasets.
Finally, AML has a better accuracy-time trade-off than baselines, which can be concluded from  Figure~\ref{fig:result}. For example, we can find that AML is faster than baselines when achieving the same accuracy. 

\subsection{Comparison with non-GNN and Local GNN-LP Baselines}

\begin{table*}[!t]
	\centering
	\scriptsize
	\caption{Comparison with non-GNN and local GNN-LP baselines. ``(S)'' in AML means using SAGE as the GNN model, and ``(G)'' in AML means using GAT as the GNN model. Accuracy of non-GNN and local GNN-LP baselines on ogbl-ppa and ogbl-citation2 is from~\protect\cite{DBLP:conf/nips/ZhangLXWJ21}. ``*'' denotes the results achieved by rerunning the authors' code on the clean ogbl-collab.}
	\label{tb:others}
	\begin{tabular}{llcccccccccc}
		\toprule
		\multirow{2}*{Category}&\multirow{2}*{Methods} &\multicolumn{2}{c}{ogbl-collab} &\multicolumn{2}{c}{ogbl-ppa} &\multicolumn{2}{c}{ogbl-citation2}\\
		\cmidrule{3-8}
		&&Hits@50(\%)$\uparrow$ &Time(s)$\downarrow$ &Hits@100(\%)$\uparrow$ &Time(s)$\downarrow$ &MRR(\%)$\uparrow$ &Time(s)$\downarrow$\\
		\midrule
		\multirow{5}*{Non-GNN}
		&CN & 49.96$\pm$0.00* & - & 27.60$\pm$0.00 & - & 51.47$\pm$0.00 & -\\
		&AA & 56.49$\pm$0.00* & - & 32.45$\pm$0.00 & - & 51.89$\pm$0.00 & -\\
		&Node2vec & 49.29$\pm$0.64* & -& 22.26$\pm$0.88 & - & 61.41$\pm$0.11 & -\\
		&MF & 37.93$\pm$0.76* & - & 32.29$\pm$0.94 & - & 51.86$\pm$4.43 & -\\
		\midrule
		\multirow{2}*{Local GNN-LP}
		&DE-GNN & \textbf{57.87$\bm{\pm}$0.79}* & $6.1\times10^4$ &45.70$\pm$3.46 & $2.0\times10^6$ & 78.85$\pm$0.17 & $1.1\times10^6$ \\
		& SEAL   & 57.55$\pm$0.72* & $6.5\times10^4$ &48.80$\pm$3.16 & $2.0\times10^6$ & \textbf{87.67$\bm{\pm}$0.32} & $1.1\times10^6$\\
		\midrule
		\multirow{2}*{Global GNN-LP}
		&AML (S) & 57.26$\pm$1.25 & $\bm{4.4\times10^3}$ & 49.73$\pm$0.89 & $\bm{3.0\times10^4}$ & 86.55$\pm$0.06& $\bm{1.0\times10^4}$ \\
		&AML (G) & 57.60$\pm$0.71 & $4.5\times10^3$ & \textbf{50.23$\bm{\pm}$0.78} & $3.2\times10^4$ & 86.70$\pm$0.05 & $\bm{1.0\times10^4}$\\
		\bottomrule
	\end{tabular}
\end{table*}

Comparison with non-GNN and local GNN-LP baselines is shown in Table~\ref{tb:others}. According to the results, we can draw the following conclusions. Firstly, AML is comparable with state-of-the-art local GNN-LP methods in accuracy. For example, AML has almost no accuracy loss compared to the best baseline DE-GNN on obgl-collab. On ogbl-ppa, AML achieves an accuracy gain of $3.43\%$  compared to the best baseline SEAL. On ogbl-citation2, AML gets an accuracy loss less than $1\%$ compared to the best baseline SEAL.
Secondly, AML is about $13{\times}{\sim}110{\times}$ faster than local \mbox{GNN-LP} baselines. For example, AML is about $13{\times}$ faster than \mbox{DE-GNN} and SEAL on ogbl-collab, about $66{\times}$ faster on ogbl-ppa, and about $110{\times}$ faster on ogbl-citation2. In particular, \mbox{DE-GNN} and SEAL need 3.5 days to get the mean accuracy on ogbl-collab which is a relatively small-scale dataset. By contrast, AML only needs 1.2 hours to get the mean accuracy. Finally, GNN-LP methods can achieve better accuracy than non-GNN methods. For example, DE-GNN improves by $13\%$ in accuracy over MF on ogbl-ppa and improves by $17\%$ over Node2vec on ogbl-citation2. AML improves by $17\%$ in accuracy over MF on ogbl-ppa and improves by $16\%$ over Node2vec on ogbl-citation2. 


\begin{table*}[!t]
	\centering
	\caption{Experiment to verify the necessity of GNN in AML. SMLP applies MLP with pre-encoding to learn representation for both head nodes and tail nodes. ``Gap'' denotes the accuracy of AML minus that of SMLP.}
	\label{tb:slpe}
	\begin{tabular}{lccccccccccccc}
		\toprule
		\multirow{2}*{Methods} &\multicolumn{2}{c}{ogbl-collab} &\multicolumn{2}{c}{ogbl-ppa} &\multicolumn{2}{c}{ogbl-citation2}\\
		\cmidrule{2-7}
		& Hits@50 (\%) $\uparrow$ & Gap & Hits@100 (\%) $\uparrow$ & Gap & MRR (\%) $\uparrow$ & Gap\\
		\midrule
		SMLP & 47.25 $\pm$ 0.89 & - &47.42 $\pm$ 1.37 & - & 69.82 $\pm$ 0.05 &-\\
		\midrule
		AML (S) & 57.26 $\pm$ 1.25 & $+10.01$ & 49.73 $\pm$ 0.89 & $+2.31$ & 86.55 $\pm$ 0.06 & $+16.73$ \\ 
		AML (G) & \textbf{57.60 $\pm$ 0.71} & $+10.35$ & \textbf{50.23 $\pm$ 0.78} & $+2.81$ & \textbf{86.70 $\pm$ 0.05} & $+16.88$\\ 
		\bottomrule
	\end{tabular}
\end{table*}

\subsection{Necessity of GNN in AML}
Here we perform experiment to verify the necessity of GNN in AML. We design a method called symmetric MLP~(SMLP) which applies MLP with pre-encoding to learn representation for both head nodes and tail nodes. More specifically, SMLP adopts the techniques for learning tail node representation in AML, i.e., (\ref{eq:prepro}) and~(\ref{eq:tail_mlp1}), to learn representation for both head and tail nodes.

Results are shown in Table~\ref{tb:slpe}. We can find that SMLP is much worse than AML on all datasets. This shows that training a GNN model for node representation learning is necessary to achieve high accuracy.

\begin{table*}[!t]
	\centering
	\caption{Reversed asymmetric learning. AML-R denotes the reversed case of AML, which learns representation for head nodes with an MLP model while learning representation for tail nodes with a GNN model. ``(S)'' means using SAGE as the GNN model, and ``(G)'' means using GAT as the GNN model.}
	\label{tb:reversed-aml}
	\begin{tabular}{lccccccccccccc}
		\toprule
		\multirow{2}*{Methods} &ogbl-collab &ogbl-ppa &ogbl-citation2\\
		\cmidrule{2-4}
		& Hits@50 (\%) $\uparrow$ & Hits@100 (\%) $\uparrow$ & MRR (\%) $\uparrow$\\
		\midrule
		AML-R (S) & 57.15 $\pm$ 0.32& 49.73 $\pm$ 0.45 & 85.70 $\pm$ 0.10\\ 
		AML (S) & 57.26 $\pm$ 1.25 & 49.73 $\pm$ 0.89 & 86.55 $\pm$ 0.06\\ 
		AML-R (G) & 57.08 $\pm$ 1.19& \textbf{50.30 $\pm$ 0.61}& 85.91 $\pm$ 0.04\\ 
		AML (G) & \textbf{57.60 $\pm$ 0.71} & 50.23 $\pm$ 0.78& \textbf{86.70 $\pm$ 0.05}\\ 
		\bottomrule
	\end{tabular}
\end{table*}
\subsection{Reversed Asymmetric Learning}
In above experiments, AML learns representation for head nodes with a GNN model while learning representation for tail nodes with an MLP model. Here we verify whether the reversed case can also behave well. We denote the reversed case as AML-R, which learns representation for head nodes with an MLP model while learning representation for tail nodes with a GNN model. Results are shown in Table~\ref{tb:reversed-aml}. Results show that \mbox{AML} and AML-R have similar accuracy.

\begin{table*}[!h]
	\centering
	\caption{Ablation study. ``KT'' represents knowledge transfer. ``$\Delta^{(L)}$'' indicates the residual variable defined in~(6). ``PE'' means pre-encoded information of graph structure. ``HO'' represents homophily. ``Gap'' denotes the accuracy of variants of AML minus that of AML.}
	\label{tb:ablation}
	\begin{subtable}{\linewidth}
		\centering
		\caption{SAGE as the base GNN model.}
		\label{tb:ablation-sage}
		\begin{tabular}{lccccccccc}
			\toprule
			\multirow{2}*{Methods} &\multicolumn{2}{c}{ogbl-collab} &\multicolumn{2}{c}{ogbl-ppa} &\multicolumn{2}{c}{ogbl-citation2}\\
			\cmidrule{2-7}
			& Hits@50 (\%) $\uparrow$ & Gap & Hits@100 (\%) $\uparrow$ & Gap & MRR (\%) $\uparrow$ & Gap\\
			\midrule
			AML (w/o KT) & 54.50$\pm$0.90 &$-2.76$ & 49.16$\pm$1.07 &$-0.57$ & 86.24$\pm$0.05 &$-0.31$\\
			AML (w/o $\Delta^{(L)}$) & 54.61$\pm$0.62 &$-2.65$ & 49.00$\pm$0.26 &$-0.73$& 86.26$\pm$0.02 &$-0.29$\\
			AML (w/o HO) & 55.70$\pm$1.75 &$-1.56$ & 47.74$\pm$0.66 &$-1.99$ & 85.75$\pm$0.10 &$-0.80$\\
			AML (w/o PE) & 43.90$\pm$0.78 &$-13.36$ & 1.97$\pm$0.27 &$-47.76$ & 60.82$\pm$0.14 &$-25.73$\\
			\midrule
			AML & \textbf{57.26$\bm{\pm}$1.25}&-&\textbf{49.73$\bm{\pm}$0.89} &-& \textbf{86.55$\bm{\pm}$0.06}&-\\ 
			\bottomrule
		\end{tabular}
	\end{subtable}
	\vskip 0.1in
	\begin{subtable}{\linewidth}		
		\centering
		\caption{GAT as the base GNN model.}
		\label{tb:ablation-gat}
		\begin{tabular}{lccccccccc}
			\toprule
			\multirow{2}*{Methods} &\multicolumn{2}{c}{ogbl-collab} &\multicolumn{2}{c}{ogbl-ppa} &\multicolumn{2}{c}{ogbl-citation2}\\
			\cmidrule{2-7}
			& Hits@50 (\%) $\uparrow$ &Gap & Hits@100 (\%) $\uparrow$ &Gap & MRR (\%) $\uparrow$&Gap\\
			\midrule
			AML (w/o KT) & 54.60$\pm$0.60 &$-3.00$ & 47.38$\pm$0.28 &$-2.85$ & 86.28$\pm$0.05 &$-0.42$\\
			AML (w/o $\Delta^{(L)}$) & 54.56$\pm$1.43 &$-3.04$ & 47.04$\pm$1.03 &$-3.19$ & 86.30$\pm$0.01 &$-0.40$\\
			AML (w/o HO) & 55.73$\pm$0.34 &$-1.87$ & 47.86$\pm$1.28 &$-2.37$ & 85.83$\pm$0.04 &$-0.87$ \\
			AML (w/o PE) & 43.95$\pm$1.01 &$-13.65$ & 2.95$\pm$0.35 &$-47.28$ & 60.89$\pm$0.22 &$-25.81$\\
			\midrule
			AML & \textbf{57.60$\bm{\pm}$0.71} & - & \textbf{50.23$\bm{\pm}$0.78} &-& \textbf{86.70$\bm{\pm}$0.05}&-\\ 
			\bottomrule
		\end{tabular}
	\end{subtable}
\end{table*}
\subsection{Ablation Study}
In this subsection, we study the effectiveness of different components in AML, including knowledge transfer, residual term $\Delta^{(L)}$, pre-encoding  graph structure and modeling the homophily. Results are shown in Table~\ref{tb:ablation}. We can find the following phenomenons. Firstly, knowledge transfer between head and tail nodes effectively improves the accuracy of AML. For example, knowedge transfer can improve the accuracy of AML by $3.00\%$ on ogbl-collab, by $2.85\%$ on ogbl-ppa and by $0.42\%$ on ogbl-citation2. Secondly, $\Delta^{(L)}$ is beneficial for AML. For example, including $\Delta^{(L)}$ in AML can improve accuracy by $3.04\%$ on ogbl-collab, by $3.19\%$ on ogbl-ppa and by $0.40\%$ on ogbl-citation2. Thirdly, modeling homophily is helpful for AML. For example, modeling homophily in AML can improve accuracy by $1.87\%$ on ogbl-collab, by $2.37\%$ on ogbl-ppa and by $0.87\%$ on ogbl-citation2. Finally, pre-encoding graph structure plays a crucial role in AML. For example, \mbox{AML} has an accuracy loss of about $13\%$ on ogbl-collab, $47\%$ on ogbl-ppa and $25\%$ on ogbl-citation2 without pre-encoding graph structure. 

\section{Conclusions}
Graph neural network based link prediction~(GNN-LP) methods have achieved better accuracy than non-GNN based link prediction methods, but suffer from scalability problem for large-scale graphs. Our computation complexity analysis reveals that the scalability problem of existing GNN-LP methods stems from their symmetric learning strategy for node representation learning. Motivated by this finding, we propose a novel method called \mbox{AML} for GNN-LP. To the best of our knowledge, \mbox{AML} is the first GNN-LP method adopting an asymmetric learning strategy for node representation learning. Extensive experiments show that \mbox{AML} is significantly faster than baselines with a symmetric learning strategy while having almost no accuracy loss.

\bibliographystyle{alpha}
\bibliography{sample}

\end{document}